\documentclass[10pt,twocolumn,letterpaper]{article}
\usepackage{authblk}
\usepackage{xspace}
\makeatletter
\DeclareRobustCommand\onedot{\futurelet\@let@token\@onedot}
\def\@onedot{\ifx\@let@token.\else.\null\fi\xspace}

\def\ie{\emph{i.e}\onedot}

\def\etal{\emph{et al}\onedot}
\makeatother
\usepackage{graphicx}
\usepackage{amsmath}
\usepackage{amssymb}
\usepackage{booktabs}
\usepackage{multirow} 
\usepackage{bbm}
\usepackage{caption}
\usepackage{subcaption}
\usepackage{soul}

\usepackage[pagebackref,breaklinks,colorlinks]{hyperref}

\usepackage[capitalize]{cleveref}
\crefname{section}{Sec.}{Secs.}
\Crefname{section}{Section}{Sections}
\Crefname{table}{Table}{Tables}
\crefname{table}{Tab.}{Tabs.}

\begin{document}

\date{}

\title{Deep Imbalanced Regression via Hierarchical Classification Adjustment}

\author[1]{Haipeng Xiong}
\author[1]{Angela Yao}
\affil[1]{National University of Singapore}
\maketitle

\begin{abstract}
Regression tasks in computer vision, such as age estimation or counting, are often formulated into classification by quantizing the target space into classes.  Yet real-world data is often imbalanced -- the majority of training samples lie in a head range of target values, while a minority of samples span a usually larger tail range.  By selecting the class quantization, one can adjust imbalanced regression targets into balanced classification outputs, though there are trade-offs in balancing classification accuracy and quantization error.  To improve regression performance over the entire range of data, we propose to construct hierarchical classifiers for solving imbalanced regression tasks. The fine-grained classifiers limit the quantization error while being modulated by the coarse predictions to ensure high accuracy. Standard hierarchical classification approaches, however, when applied to the regression problem, fail to ensure that predicted ranges remain consistent across the hierarchy. As such, we propose a range-preserving distillation process that can effectively learn a single classifier from the set of hierarchical classifiers. Our novel hierarchical classification adjustment (HCA) for imbalanced regression shows superior results on three diverse tasks: age estimation, crowd counting and depth estimation. We will release the source code upon acceptance.
\end{abstract}

\begin{figure}[!h]
\centering
\includegraphics[width=0.45\textwidth]{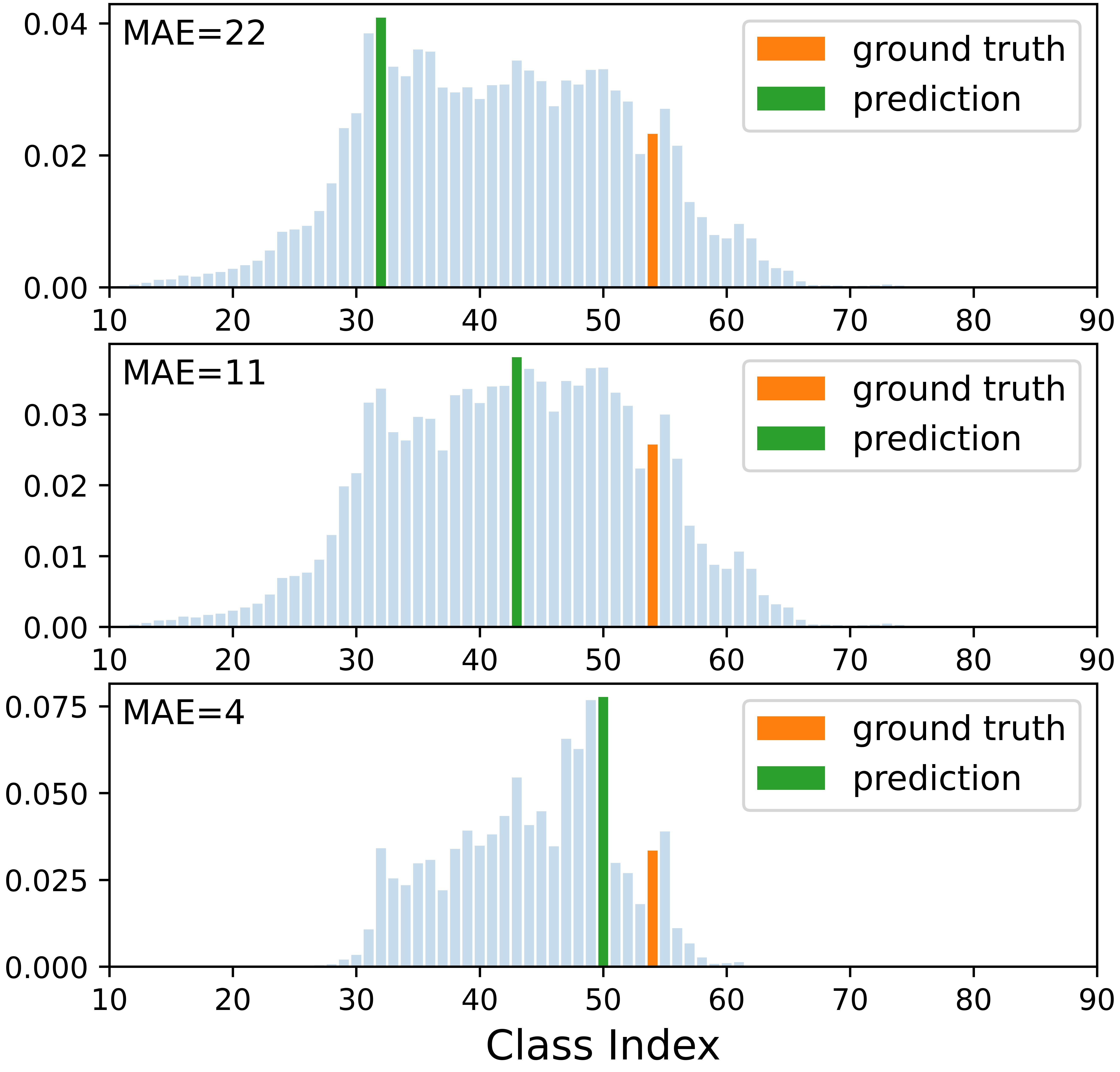}
\caption{Hierarchical classification adjustment (HCA). At the top is the finest classifier $H$.  Progressing from the top to the middle and bottom, classifier $H$ is adjusted by coarser classifiers ($1\sim H-1$) which bring the prediction closer to the ground truth. 
}
\label{fig:hca_motivation}
\end{figure}

\section{Introduction}

In machine learning, classification is used to predict categorical targets, while regression is used for predicting continuous targets.  Yet for many regression tasks in computer vision, such as depth estimation~\cite{cao2017estimating}, age estimation~\cite{rothe2015dex} and crowd-counting~\cite{ liu2019counting}, performance is better under a classification formulation.   
Most commonly, the continuous output is quantized, and each bin is treated as a class.  

Depending on the data distribution of the output space, different quantization schemes are more suitable, \ie, linear~\cite{xiong2019open} if the data is approximately balanced, versus  logarithmic~\cite{fu2018deep} or adaptive~\cite{wang2021uniformity} for long-tail and imbalanced data.  An adaptive quantization can transform imbalanced regression targets into more balanced class labels.  The number of classes and the target range each class covers should be selected to ensure a sufficient number of samples for learning.  However, the larger the interval, \ie to ensure balanced and sufficient class samples, the greater the quantization error when recovering the target regression values.  In practice, the number of classes and quantization scheme is chosen to trade off the classification accuracy and quantization error~\cite{cao2017estimating,xiong2022discrete}.

By transforming regression to classification, can we apply imbalanced classification approaches? Existing long-tail classification works try to balance the input samples~\cite{byrd2019effect,yu2022re}, features~\cite{park2022majority,kang2021exploring,hong2022safa}, losses~\cite{NIPS2017_147ebe63,sinha2020class} or output logits~\cite{menon2021longtail,zhong2021improving,li2022long} during training. It is challenging to re-balance a single classifier to cover both the head and the tail; often, improvements on the tail classes come at the expense of harming the head. 

To capture the entire target range, one alternative is to ensemble a set of classifiers~\cite{liu2008exploratory,hido2009roughly,li2022nested}. 
For regression converted to classification settings, a convenient way to create an ensemble of diverse classifiers is to apply different quantizations. 
When merging the ensemble, we wish to benefit from the higher performance of coarse classifiers while preserving the resolution of fine classifiers.  

In this work, we advocate for the \emph{adjustment} of fine-grained classifiers with progressively coarser ones. We refer to this procedure as Hierarchical Classification Adjustment, or HCA. HCA works with the logits of the classifier ensemble; as shown in Fig.~\ref{fig:hca_motivation}, adding the coarser (but more accurate) logits progressively improves the accuracy. 

\begin{figure}[!t]
\centering
\includegraphics[width=0.45\textwidth]{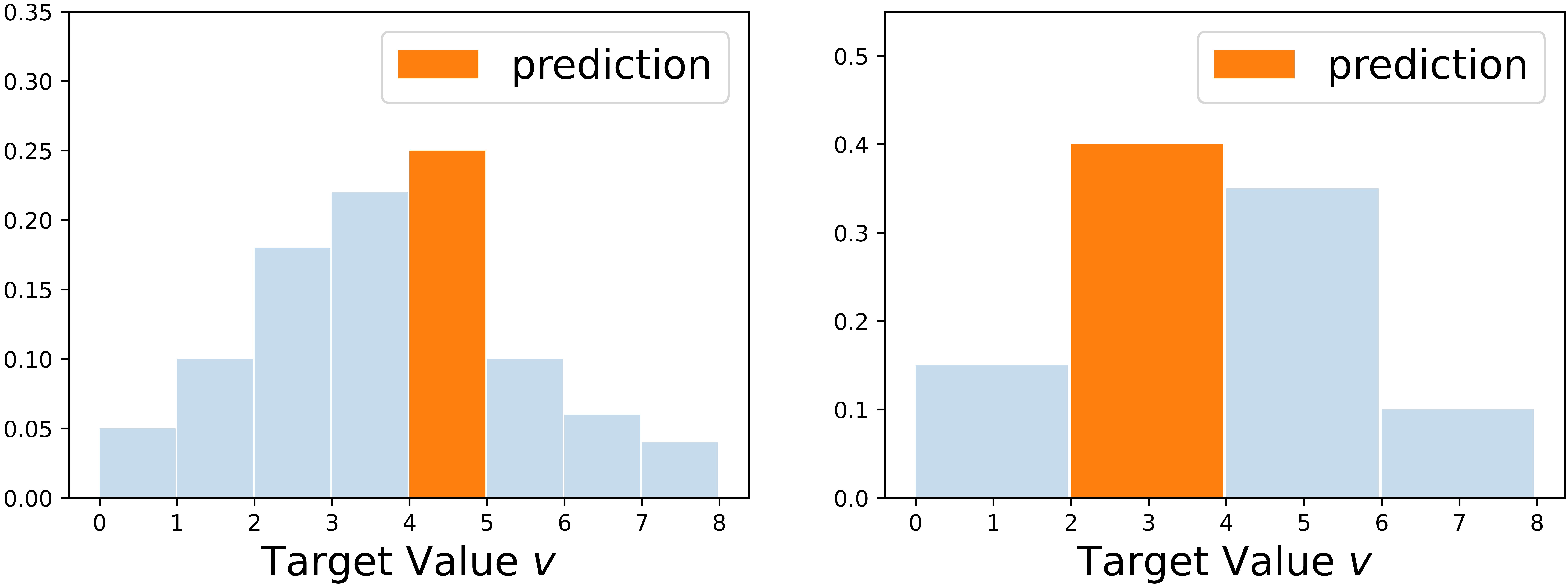}
\caption{ (Left) The prediction of an 8-class classifier. (Right) Downsampling logits with eq.~\eqref{eq:sum_pool} 
fails to preserve the range.  }
\label{fig:hd_inconsist}
\end{figure}

In addition, we propose to distill the entire hierarchical ensemble into a single classifier; we refer to this process as HCA-d. 
HCA-d is inspired by~\cite{bertinetto2020making,garg2022learning}, which distill image classifiers from a set of classifier learned with multi-granular image labels.  However, such approaches applied to the regression setting are not range-preserving.  Specifically, across the classifiers in the ensemble, the estimated target ranges do not remain consistent (see detailed example in Fig.~\ref{fig:hd_inconsist}). To mitigate the errors that may arise, we propose a range-preserving adjustment to ensure that HCA-d remains consistent in the distillation process.  Overall, HCA-d is simple but efficient, showing improvements over the whole range of the target space. Our contributions can be summarized as:

\begin{itemize}
    \item A novel Hierarchical Classification Adjustment (HCA) 
    to merge a finely-quantized classifier with an ensemble of progressively coarser classifiers over an imbalanced target range; 
    \item A range-preserving distillation technique, HCA-d, which ensures 
    consistent class (range) predictions  predicted 
    across the hierarchy of classifiers. 
    \item HCA shows comparable or superior 
    performance on imbalanced visual regression tasks, including age estimation, crowd counting and depth estimation. 
\end{itemize}

\section{Related Works}

Real-world data~\cite{silberman2012indoor,MCNN_2016_CVPR} is commonly imbalanced 
so standard approaches try to re-weight the importance between head and tail samples~\cite{khoshgoftaar2010comparing,menon2021longtail}. 
Most works~\cite{byrd2019effect,yu2022re,park2022majority,kang2021exploring,hong2022safa,NIPS2017_147ebe63,sinha2020class,liu2008exploratory,hido2009roughly,li2022nested} focus on imbalanced classification, while only a few~\cite{yang2021delving,ranksim_icml2022,steininger2021density,ren2022bmse} studied imbalanced settings in a regression.\\

\noindent\textbf{Imbalanced Classification} Current research working with single classifiers tries to improve tail class performance via sample weighting~\cite{byrd2019effect,yu2022re,sinha2020class}, upsampling~\cite{park2022majority,hong2022safa} or adjusting class margins~\cite{menon2021longtail,zhong2021improving,li2022long}. 
However, a trade-off is hard to achieve with a single classifier and invariably sacrifices the head for the tail. 

While a single classifier may not be suitable, ensembling multiple classifiers can {cover all the classes}~\cite{hido2009roughly,khoshgoftaar2010comparing,wang2021longtailed,li2022nested,xu2022constructing}. In an ensemble, a critical issue is ensuring individual learners' diversity. A classic approach to introduce diversity is bagging~\cite{breiman1996bagging} - sampling with replacements for different training data partitions for each of the classifiers. Hido \etal~\cite{hido2009roughly} sample 
subsets, which are individually imbalanced but globally balanced when averaging all subsets, to ensure balance and diversity. Li \etal~\cite{li2022nested} construct individual learners by hard example mining. Xu \etal~\cite{xu2022constructing} progressively split head and tail classes to alleviate imbalance and learn diverse classifiers. 
For deep imbalanced regression, it is convenient to get diverse classifiers by applying different quantization strategies without splitting the dataset. 

\noindent\textbf{From Imbalanced Classification to Regression}
Imbalanced regression~\cite{yang2021delving,ranksim_icml2022,steininger2021density,ren2022bmse} is less explored than classification. 
Most works~\cite{yang2021delving,steininger2021density,ren2022bmse} are inspired by classification techniques. 
Yang~\etal~\cite{yang2021delving}
showed that regression values are locally correlated and can benefit from a smoothed version of sample weighting. 
Ren~\etal~\cite{ren2022bmse} is 
inspired from logit adjustment~\cite{menon2021longtail} and proposed a balanced Mean Square Error (BMSE) loss. Conversely, Liu~\etal~\cite{liu2019counting} and Wang~\etal~\cite{wang2021uniformity} choose distribution-aware quantization to transform imbalanced regression into a less imbalanced classification problem. 
We also transform imbalanced regression into classification but explore hierarchical classifiers, where each classifier trades off head and tail performances differently while the combination is suitable to the whole ranges.

\noindent \textbf{Hierarchical classification}~\cite{bertinetto2020making,chang2021your,garg2022learning,wu2016learning} leverages the taxonomy or hierarchical structure of classes to ensure more semantically meaningful mistakes. 
For example, a poodle is better to be mis-classified as a dog instead of a cat. To learn the hierarchy, hierarchical classifiers are trained together in~\cite{bertinetto2020making}. A key issue is how to align hierarchical outputs and propagate supervision from the coarse to the fine classifiers~\cite{bertinetto2020making,chang2021your,garg2022learning,wu2016learning}. 
The standard approaches~\cite{bertinetto2020making,garg2022learning} treat classifier outputs after the softmax as posterior probabilities and sum them. 
Such a paradigm does not ensure consistent 
predictions when pooling fine-grained predictions to a coarser level. 
For regression, such inconsistencies adversely affect the learning and serve as the motivation for our proposed range-preserving distillation.

\begin{figure*}[!h]
\centering
 \begin{subfigure}[b]{0.23\textwidth}
     \centering
     \includegraphics[width=\textwidth]{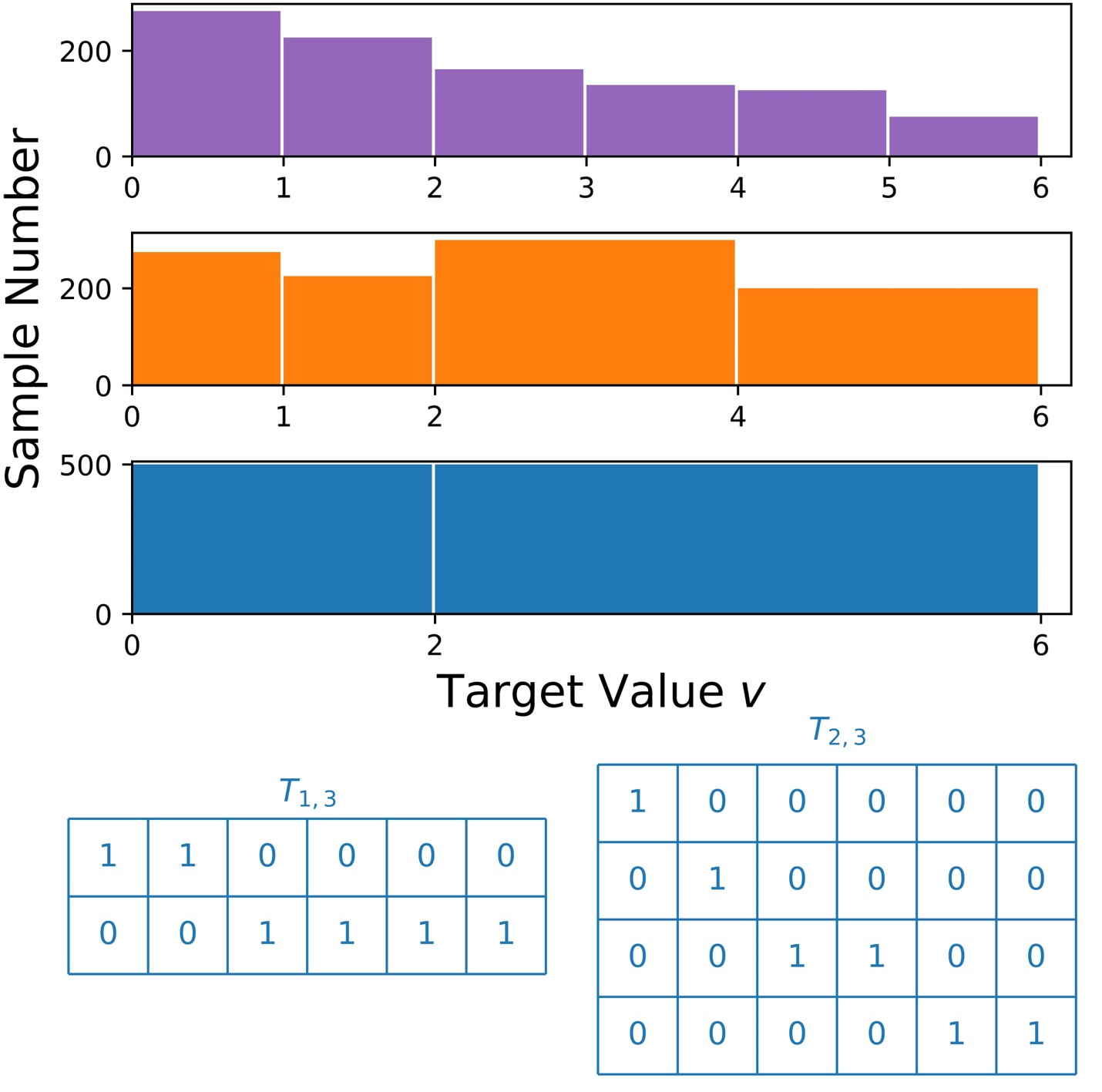}
     \caption{}
 \end{subfigure}
\hfill
 \begin{subfigure}[b]{0.38\textwidth} 
     \includegraphics[width=\textwidth]{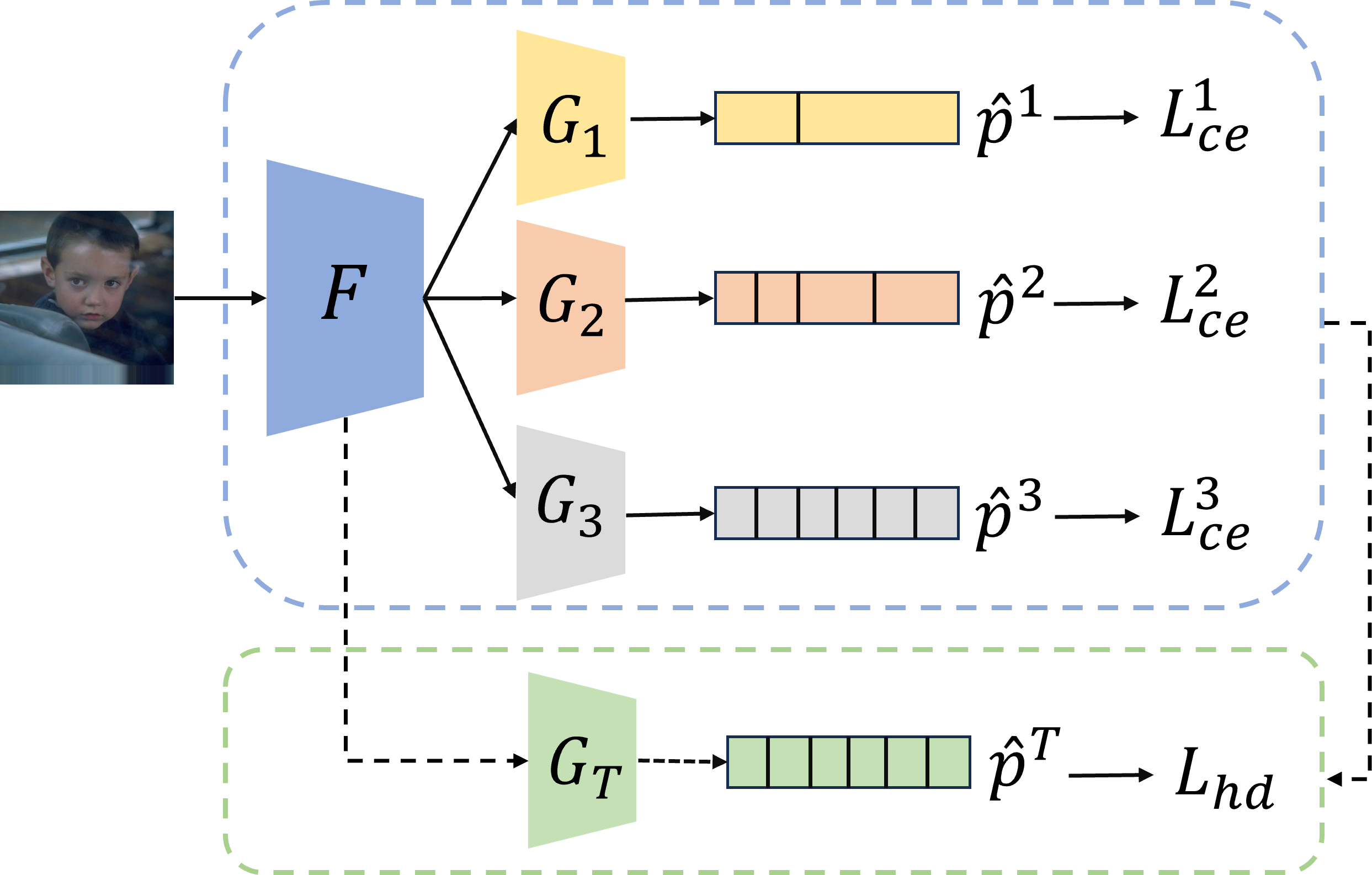}
     \caption{}
 \end{subfigure}
\hfill
 \begin{subfigure}[b]{0.32\textwidth} 
     \includegraphics[width=\textwidth]{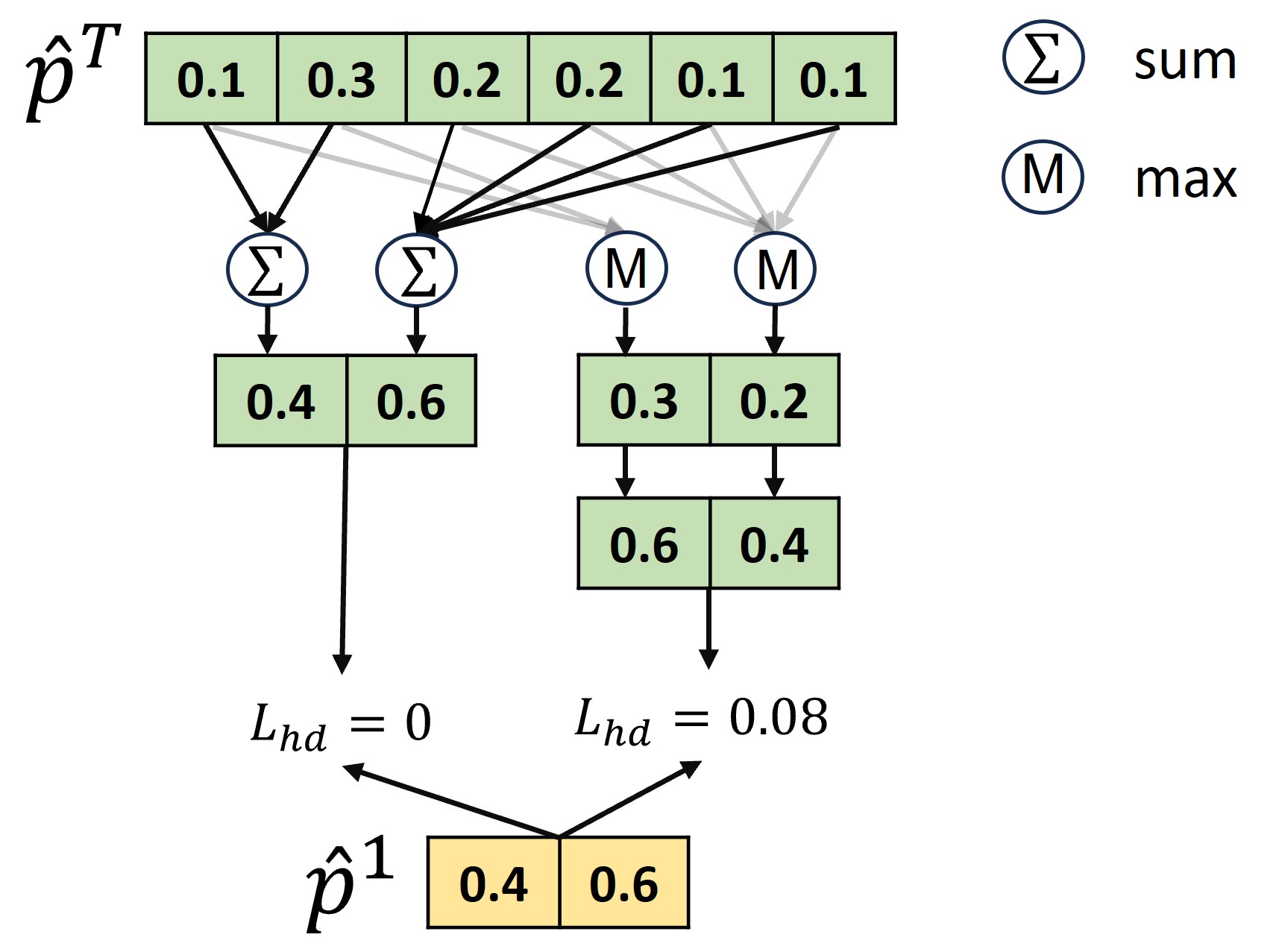}
     \caption{}
 \end{subfigure}

\caption{An example of $H=3$ hierarchical classifiers. (a) Class splitting and label distributions for hierarchical classifiers $1 \sim 3$ from bottom to top. We also visualize class transition matrices $T_{h,H}$ between the $h$-th and $H$-th classifiers, which are used to merge hierarchical predictions in eq.~\eqref{eq:add_comb} and eq.~\eqref{eq:mul_comb}. (b) Training of Hierarchical classifiers 
(dashed blue box). The dashed green box denotes the second-stage learning, which distillates classifier $T$ from hierarchical predictions $\hat{p}^{h}$. (c) Distillation learning of classifier $T$. Hierarchical alignment with eq.~\eqref{eq:sum_pool} and eq.~\eqref{eq:max_pool} are compared. Here we only plot the distillation from $\hat{p}^1$, while from $\hat{p}^2$ and $\hat{p}^3$ are not visualized. }
\label{fig:hier_example}
\end{figure*}

\section{Hierarchical Classification Adjustment}

We propose a Hierarchical Classification Adjustment (HCA) for imbalanced regression. It learns an ensemble of hierarchical classifiers 
and then leverages the set of 
predictions to improve the performance of few-shot ranges while maintaining the performance of many-shot ranges. In this section, we first describe how to set a single classifier for a regression problem (Sec.~\ref{subsec:plain_class}), then partition the continuous label space into an ensemble of discrete hierarchical classes (Sec.~\ref{subsec:hier_generate}), followed by the hierarchical adjustment (Sec.~\ref{subsec:HCA}) and distillation (Sec.~\ref{subsec:HCAdist}). 

\subsection{A Vanilla Classifier for Continuous Targets}\label{subsec:plain_class}
Consider a continuous dataset $D=\{x,v\}$, where $x$ and $v$ denote the input and target value, respectively. Let $V_{min}$ and $V_{max}$ denote the minimal and maximal values of $v$ in the training set. Like~\cite{xiong2019open,liu2019counting}, we divide the target range $[V_{min}, V_{max}]$ into $C$ intervals  $(V_{min},V_1]$, $(V_1,V_2]$, $...$, $(V_{C-1},V_{max}]$ and treat samples within each interval as samples belonging to classes $c = 1 ... C$.  A standard classifier can be trained to estimate the interval index $c$ based on feature representations of $x$. Consider an input sample $x$, represented by a feature $f \in \mathbbm{R}^{d}$ extracted by 
network $F$: 
\begin{equation}
    f = F(x),
\end{equation}
with a predicted 
class logit $\hat{p} \in \mathbbm{R}^{C_h}$ where
\begin{equation}
    \hat{p} = \text{Softmax}\{G(f)\}
\end{equation}
and G is a mapping function with learnable weights. 

For learning $F$ and $G$, we apply a cross entropy loss $L_{ce}$ with $\hat{p}$
\begin{equation} 
    L_{ce} = -\sum_{j=1}^{C} {p}[j] \times log(\hat{p}[j]).
\end{equation}
where $p \in \mathbbm{R}^{C}$ is the one-hot ground-truth. We can also apply label smoothing to  $p_i$; 
the soft ordinal loss (SORD)~\cite{diaz2019soft}
applies a Gaussian smoothing to ensure that ordinal relationships are partially preserved in the target classes. 

After training, the predicted class 
can be determined by the max dimension of $\hat{p}$. The class is then mapped back to a representative regression value for evaluation.  Following~\cite{wang2021uniformity}, we choose the mean values of samples fallen in each class interval. 
For age estimation, we adopt linear intervals with length $1$, since ages increase with step $1$; while for counting or depth estimation tasks, we choose log-spaced intervals as per~\cite{liu2019counting,xiong2022discrete} for fair comparison.

\subsection{Hierarchical Classifier Ensemble}\label{subsec:hier_generate}

Consider $H$ classifiers; these classifiers are hierarchical, in that each covers a progressively coarser quantization. 
Consider the finest quantization, which we designate as the setting of the $H$-th classifier. We can merge classes of the $H$-th classifier to form coarser quantized classifiers.  For $h=1$ to $H-1$, the classifier has $C_h = 2^h$ classes, where each class interval's range is determined to normalize the number of data samples per class.  For example, for the first classifier ($h\!=\!1$), the two classes cover ranges $(V_{min}, V_{med}]$ and $[V_{med}, V_{max})$, where $V_{med}$ is the value selected from $V_i$ that is closest to the median; for $h\!=\!2$, the 4 intervals are selected in $V_i$ to cover quartiles of the data samples. 
Fig.~\ref{fig:hier_example} (a) shows an example of $H=3$ hierarchical classifiers. We can observe that the label distribution of $1\sim (H-1)$ hierarchical classifiers is more balanced than the $H$-th classifier.

The $h$-th classifier predicts $\hat{p}^{h} \in \mathbbm{R}^{C_h}$ based on
\begin{equation}
     \hat{p}^{h} = \text{Softmax}\{G_h(f)\},
\end{equation}
where $G_h$ is a mapping function with learnable weights. 

Its cross entropy $L_{ce}^h$ can be given as 
\begin{equation} \label{eq:ce_single}
    L_{ce}^h = -\sum_{j=1}^{C_h} {p}^{h}[j] \times log(\hat{p}^{h}[j]),
\end{equation}
where $p \in \mathbbm{R}^{C_h}$ is the ground-truth for the $h$-th classifier.

The overall loss for training feature network $F$ and hierarchical classifiers $G_h$ is the sum of all the cross-entropies:

\begin{equation} \label{eq:ce_loss}
    L=\sum_{h=1}^{H} L_{ce}^{h}.
\end{equation}
Note that we do not weight each $L_{ce}^{h}$ differently since they have the same scale.

\subsection{Hierarchical Classifier Adjustment (HCA)}\label{subsec:HCA}
In the ensemble of classifiers learned by Eq.~\eqref{eq:ce_loss}, classifier $H$ has the finest quantization (and therefore the lowest quantization error) but is also the least accurate.  In contrast, as the classifier gets progressively coarser, it gets more accurate, but also has higher quantization error (see Fig.~\ref{fig:hier_example}). To merge these results, we can adjust the prediction of classifer $H$ with the coarser classifiers $H-1$ to $1$.

Specifically, from the hierarchical predictions $\hat{p}_{i}^{h}$, we can estimate an adjusted prediction 
through a summation
\begin{equation}\label{eq:add_comb}
    \hat{p}^{a} =  \hat{p}^{H} + \sum_{h=1}^{H-1}  T_{h,H}^{T} \cdot  \hat{p}^{h},
\end{equation}
or a multiplication
\begin{equation}\label{eq:mul_comb}
    \hat{p}^{m} =   log (\hat{p}^{H}) +  \sum_{h=1}^{H-1} T_{h,H}^T \cdot  log (\hat{p}^{h}),
\end{equation} 
operation.  In Eqs.~\ref{eq:add_comb} and~\ref{eq:mul_comb}, 
$p^a,p^m$ are addition- and multiplication-adjusted predictions that keep the finest quantization as $H$-th classifier. $T_{h,H}\in \mathbbm{R}^{C_h\times C_H }$ is the class mapping from $h$-th classifier to $H$-th classifier. If the $u$-th class in $H$-th classifier is the $v$-th class in the $h$-th classifier, then $T_{v,u}=1$; otherwise $T_{v,u}=0$. Fig.~\ref{fig:hier_example} (a) visualizes an example of $T_{h,H}$ for $H=3$ hierarchical classifiers. Note that the multiplication merging in Eq.~\eqref{eq:mul_comb} has a similar form as logit adjustment, but here we use hierarchical prediction $\hat{p}^h$ to adjust $\hat{p}^H$ rather than the frequency of each class.  The final class is recovered by taking a max over $p^a$ or $p^m$ for addition or multiplication adjustments respectively.   

HCA, while proposed with the concept of adjusting the finest-quantized classifier with coarser ones, is effectively an ensembling approach, voting with either the logits (Eq.~\eqref{eq:add_comb}) or log of the logits (Eq.~\eqref{eq:mul_comb}).  However, such an ensembling approach cannot ensure that the adjusted or ensembled result $p^a$ and $p^m$ will predict a final result consistent with $p^h$. 

\subsection{Range-Preserving Distillation (HCA-d)} \label{subsec:HCAdist}
In addition to the inconsistencies, the adjustment procedure, like other ensembling methods, is inefficient because it requires running $H$ classifiers during testing.  
Alternatively, we propose to distill the ensemble of classifiers into a single adjusted classifier.  The ensemble is learned during training in a first stage, frozen, and then distilled into a single classifier during a second training stage; during inference, only the adjusted classifier is applied. Such an approach is motivated by hierarchical classification~\cite{bertinetto2020making,garg2022learning}, which also distills hierarchical classifiers, though their aim is to learn hierarchy-aware features. 

Consider a classifier $T$ which predicts $\hat{p}\in R^d$ with a mapping function $G_T$:
\begin{equation} \label{eq:2-fc}
    \hat{p}^{T} = \text{Softmax} (G_T(f)),
\end{equation}

where $\hat{p}^{T}$ distills the hierarchical information of $\hat{p}^h$.  This can be achieved by adopting a Kullback–Leibler divergence loss the between softmax normalized logits $\hat{p}^{T}$ and $\hat{p}^h$. 

As $\hat{p}_{i}^{T}\in R^{C_H}$ and $\hat{p}_{i}^{h}\in R^{C_h}$ have different resolutions, they must be aligned before the distillation. 

Previous works on hierarchical classification~\cite{bertinetto2020making,garg2022learning} view $\hat{p}^{T}\in R^{C_H}$ as posterior probabilities 
and thus simply sum the corresponding dimensions in $\hat{p}^{T}$ to get a down-sampled versions of $\overline{p}^{T,h}\in R^{C_h}$ to match with $\hat{p}^h$, \ie 
\begin{equation} \label{eq:sum_pool}
    \overline{p}^{T,h} [j]= \sum_{k=1}^{C_H} T_{h,H}[j,k] \times \hat{p}^{T}[k].
\end{equation}

\noindent After aligning $\hat{p}^{T}$ with the individual $\hat{p}^h$, we can apply the Kullback–Leibler (KL) divergence between $\hat{p}_{i}^{h}$ and $\overline{p}_{i}^{T,h}$: 
\begin{equation}\label{eq:hier_h}
    L_{\text{hd}}^{h} = \text{KL}\{\hat{p}^{h}||\overline{p}^{T,h}\},
\end{equation}
and an overall hierarchical distillation by summing over all the classifiers:
\begin{equation}\label{eq:hier_all}
   L_{\text{hd}} = \sum_{h=1}^{H} L_{\text{hd}}^{h}.
\end{equation}

The hierarchical loss in Eq.~\eqref{eq:hier_h} is not range-preserving when we choose Eq.~\eqref{eq:sum_pool} as the hierarchical alignment. As indicated in Fig.~\ref{fig:hier_example} (c), $L_{hd}=0$ does not indicate the class predicted by $\hat{p}^{T}$ is within the range of classes predicted by $\hat{p}^{h}$. We can adjust Eq.~\eqref{eq:sum_pool} to range-preserving by considering the maximum of $T_{h,H}[j,k]$:

\begin{equation} \label{eq:max_pool}
    \ddot{p}^{T,h} [j]= \max_{ k=1,...,C_H} T_{h,H}[j,k] \times \hat{p}^{T}[k].
\end{equation} 
and then $\ddot{p}^{T,h}$ is normalized to get $\overline{p}^{T,h}\in R^{C_h}$ 
\begin{equation} \label{eq:max_normalize}
    \overline{p}^{T,h}[j] = \frac{\ddot{p}^{T,h}[j]}{\sum_{ l=1}^{C_h} \ddot{p}^{T,h}[l]}.
\end{equation}

\noindent \textbf{Proposition (Range-Preserving)}:  Let $v=\text{argmax}_{j} \overline{p}^{T,h}[j]$, $u=\text{argmax}_{k} \hat{p}^T[k]$.  If $\overline{p}^{T,h}$ is computed by eqs.~\eqref{eq:max_pool} and~\eqref{eq:max_normalize}, then $T_{h,H}[v,u]=1$, which indicates the class predicted by $\hat{p}^T$ is within the range of that predicted by $\overline{p}^{T,h}$.

\begin{figure*}[!h]
\centering

  \begin{subfigure}[b]{0.32\textwidth}
     \centering
     \includegraphics[width=\textwidth]{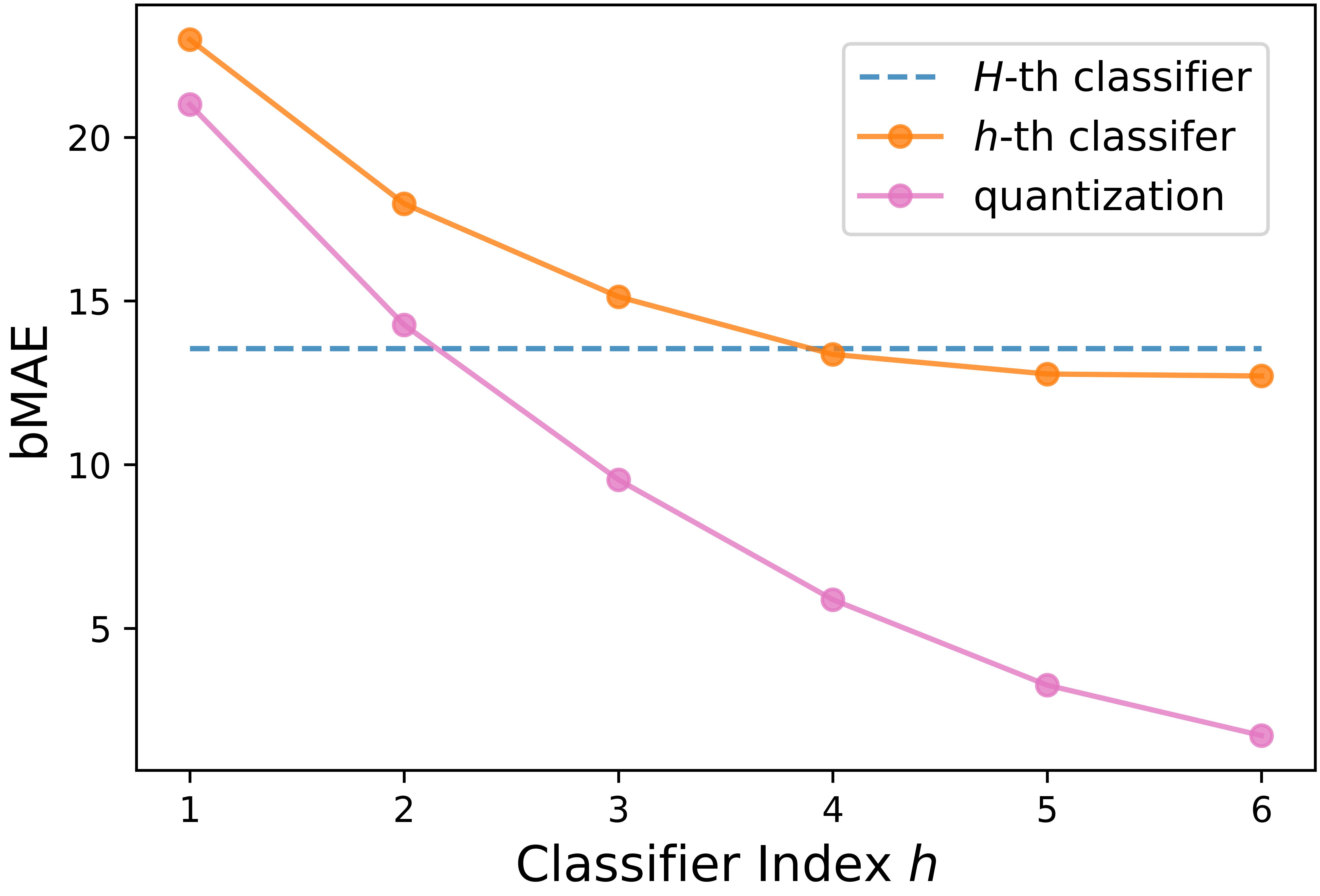}
     \caption{Quantization and Prediction Error}
     \label{fig:head_mae}
 \end{subfigure}
\hfill
 \begin{subfigure}[b]{0.32\textwidth} 
      \includegraphics[width=\textwidth]{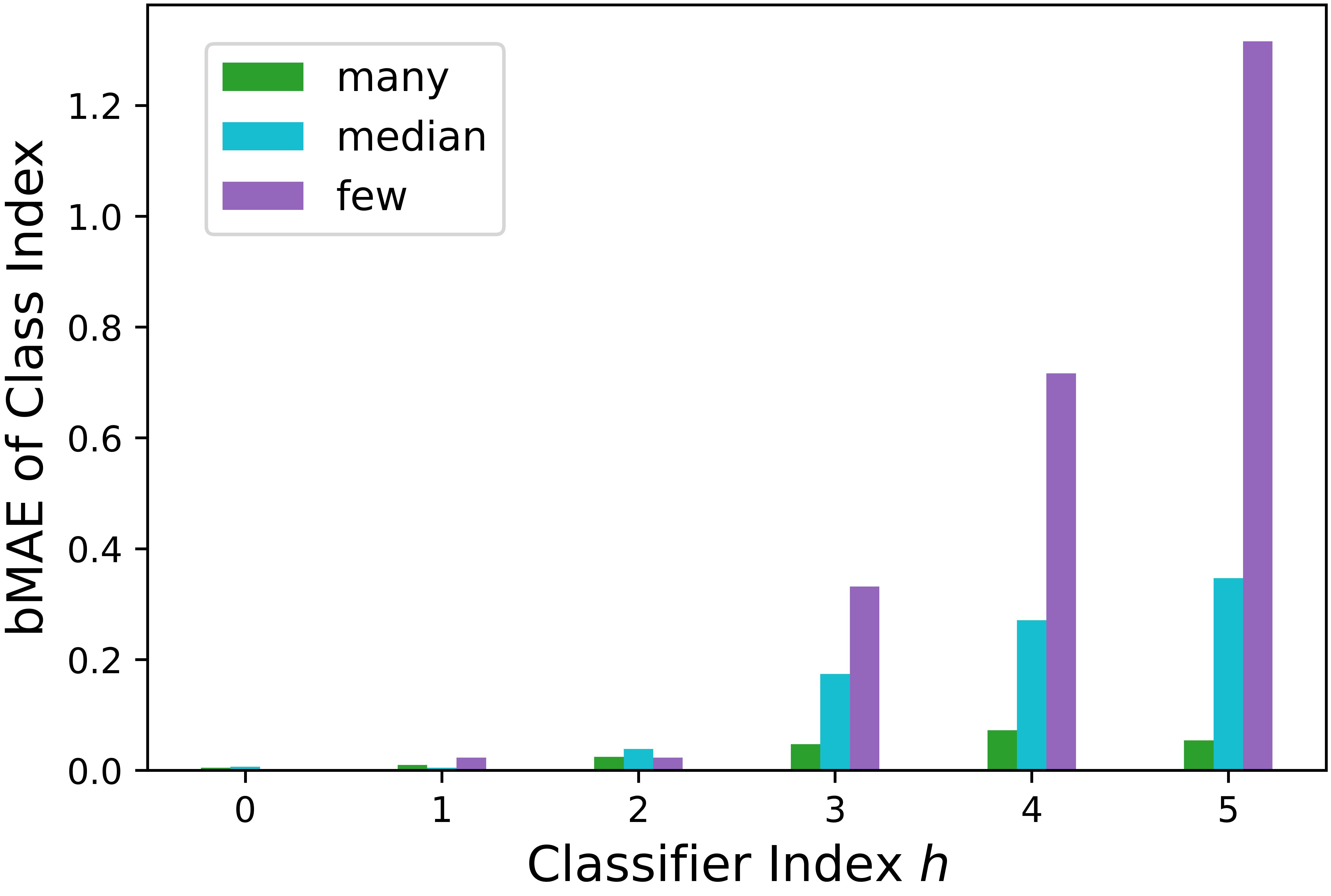}
     \caption{Individual Classifier Error}
 \end{subfigure}
\hfill
 \begin{subfigure}[b]{0.32\textwidth} 
     \includegraphics[width=\textwidth]{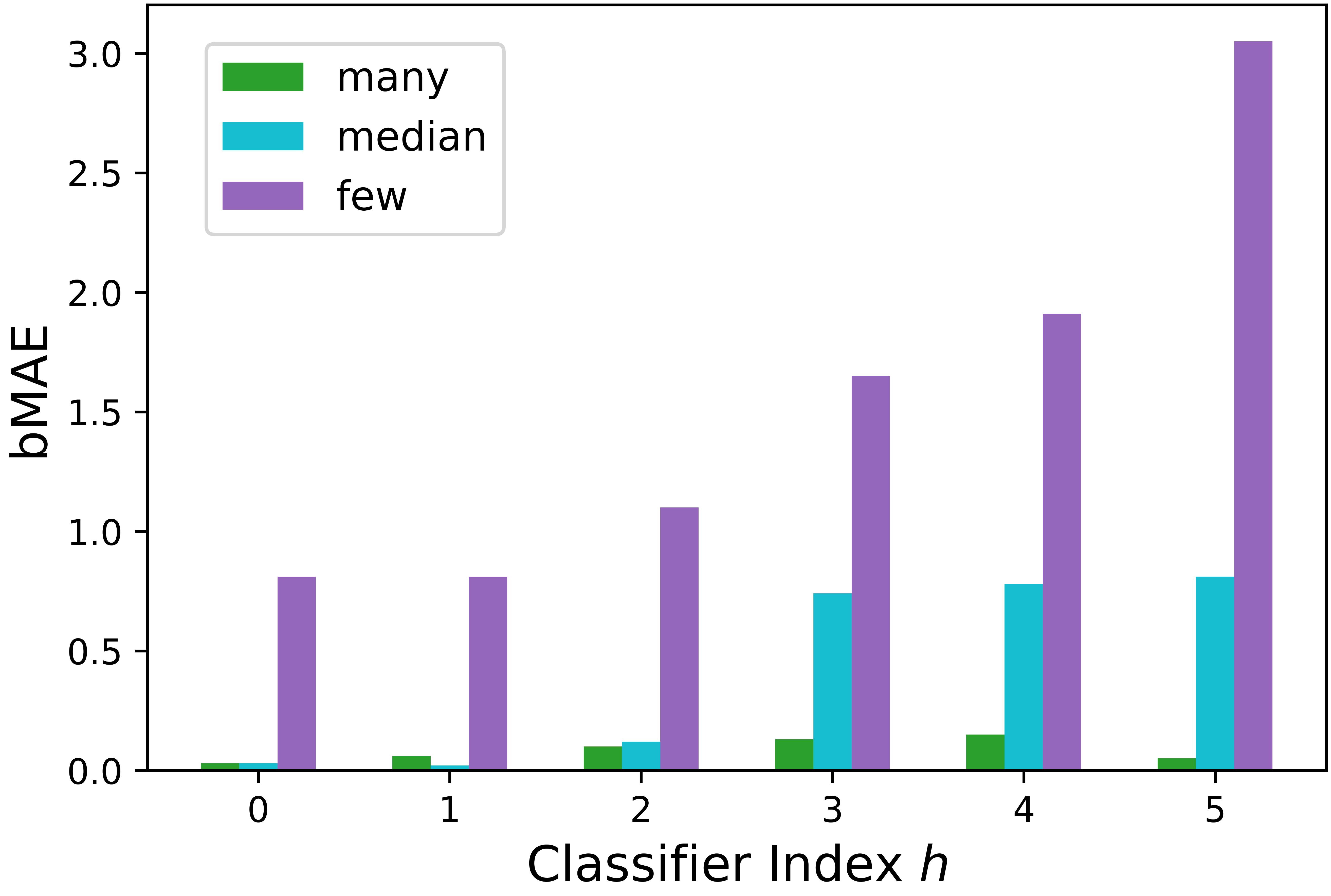}
     \caption{Adjusted Classifier Error}
     \label{fig:head_clsmae}
 \end{subfigure}
\caption{Analysis of hierarchical classifiers on IMDB-WIK-DIR dataset~\cite{yang2021delving}. (a) Comparison between quantization error and bMAE of the $h$-th classifier; for bMAE, lower is better. (b) Decrements of bMAE of the class index in each hierarchical level of classes. We report the decrements from the value of the vanilla $H$-th classifier. (c) Decrements of bMAE when adjusting the $H$-th classifier with a coarse $h$-th classifier. Specifically, a coarse $h$-th classifier provides the range, and then the finer class is selected in this range according to the prediction of the $H$-th classifier. We have subtracted the value of the vanilla $H$-th classifier. }
\label{fig:hier_analy}
\end{figure*}

\section{Experiment and Discussion}

\subsection{Implementation Details}
We conduct experiments on three imbalanced regression tasks: IMDB-WIKI-DIR~\cite{yang2021delving} for age estimation, SHTech~\cite{MCNN_2016_CVPR} for crowd counting, and NYUDv2-DIR~\cite{yang2021delving}. For the age and depth datasets, we follow the same ResNet50~\cite{he2016deep} backbone for feature extraction and training setting as~\cite{yang2021delving}. For SHTech, we use VGG16~\cite{simonyan2014very} backbone and the same training setting as~\cite{xiong2022discrete}. The finest class numbers $C_H$ are $121$ for ages and $100$ for depth and counting. Since $C_H\leq 2^7$ for all datasets, we set $H$ as 7 for all datasets. For $G_h$ $(h=1,...,H)$, we adopt one linear layer, which maps features $f\in R^{d}$ to outputs $\hat{p}^{h} \in R^{C_h}$. For $G_T$, a linear layer is also feasible, while a non-linear mapping is more adequate to distill the hierarchical information. Specifically, we adopt two fully connected layers with hidden dimensions $\frac{d}{4}$ and softplus activation. Detailed experiments of $G_T$ can be found in the supplementary.

We first train the hierarchical classifiers with the summed cross-entropy loss in Eq.~\eqref{eq:ce_loss}. We can then apply learning-free HCA results, HCA-add and HCA-mul with Eq.~\eqref{eq:add_comb} and Eq.~\eqref{eq:mul_comb}, respectively. For range-preserving HCA (HCA-d), classifiers $1\sim H$ and feature extraction network $F$ are fixed. Only classifier $T$ is trained with $L_{hd}$ in Eq.~\eqref{eq:hier_all} for additional $20\%$ epochs of stage 1 until convergence.  

\subsection{Ablation Studies}\label{subsec:ablation_study}

We first do ablation studies to verify some factors of HCA, including ground-truth labels, hierarchical class settings and two variants of HCA. IMDB-WIKI-DIR~\cite{yang2021delving} and SHTech Part A (SHA)~\cite{MCNN_2016_CVPR} datasets are chosen for ablation studies. Mean absolute error (MAE) and its balanced version bMAE~\cite{ren2022bmse} are adopted as evaluation metrics for SHA  and IMDB-WIKI-DIR, respectively. Lower MAE and bMAE denote better performance. 

\noindent $i$) \textit{One-hot or Gaussian-smoothed ground-truth labels} 
One-hot and Gaussian-smoothed~\cite{diaz2019soft} ground truths $p^h$ are two common choices for cross-entropy losses in eq.~\eqref{eq:ce_single}. Compared to one-hot $p^h$, Gaussian-smoothed ground truths further encode the ordinal relationship among labels. We compare both of them in Table~\ref{tab:hardorsoft}. From Table~\ref{tab:hardorsoft}, we can observe that HCA shows improvements with both hard and soft ground truths, and HCA with soft ground truths delivers better performance. We use soft labels by default in all of the remaining experiments.

\begin{table}[!h] \small
    \centering
    \begin{tabular}{c|c|cccc|c}
    \hline
    \multirow{2}{*}{Method}&\multirow{2}{*}{GT}& \multicolumn{4}{c|}{IMDB-WIKI-DIR}&\multirow{2}{*}{SHA}\\
    \cline{3-6}
    &&All&Many&Med.&Few&\\ 
    \hline
    CLS&\multirow{2}{*}{one-hot}&13.48&7.25&13.65&32.57&58.8\\
    HCA-d&&12.93&7.20&12.81&30.71& 55.0\\
    \hline

    CLS&\multirow{2}{*}{soft~\cite{diaz2019soft}}&13.58&7.13&	 13.95&	 33.21&58.2\\
    
    HCA-d&&12.70&	      7.00&	     13.18&	     29.94&53.7\\
    \hline

    \end{tabular}
    \caption{Soft vs. hard one-hot ground truth of classification.} 
    \label{tab:hardorsoft}
\end{table}

\noindent  \textit{$ii$) Hierarchical Class Settings} 
Combining extra classifiers could improve a single vanilla classifier, but could we just duplicate the vanilla classifier at the finest level rather than setting hierarchical classifiers? Besides, how about splitting hierarchical classifiers that equalize the interval length of each class rather than equalize sample numbers? 
We compare hierarchical class settings in Table~\ref{tab:hier_ints}. Compared with a single classifier, assembling duplicated classifiers can be helpful (overall bMAE from $13.50$ to $13.42$), but the improvement is limited compared to that of HCA. 
Moreover, it is more beneficial to split hierarchical classes by equaling the sample number within each class rather than equaling the length of class intervals.

\begin{table}[!h] \small

    \centering 
    \begin{tabular}{c|cccc|c}
    \hline
    \multirow{2}{*}{Configuration}& \multicolumn{4}{c|}{IMDB-WIKI-DIR}&\multirow{2}{*}{SHA}\\
    \cline{2-5}
    &All&Many&Med.&Few\\

    \hline
   
Single CLS&13.58&7.13&	 13.95&	 33.21&58.2\\

Same CLSs&13.42&7.10&14.38&32.22&57.9\\
\hline
E-Num HCA-d&12.70&7.00&13.18&29.94&53.7\\

E-Len HCA-d&12.77&7.23&12.92&29.77&56.2\\
    \hline
    \end{tabular}
    \caption{Comparison of various (hierarchical) class settings. ``Same CLSs'' means $H$ classifiers adopts the same class splitting as the $H$-th classifier. ``E-Num'' means equaling the number of samples within each class during hierarchical class splitting, while ``E-Len'' will equal the length of each class interval. }
    \label{tab:hier_ints} 
\end{table}

\noindent $iii$) \textit{Comparing two variants of HCA }  
Learning-free HCA and range-preserving HCA (HCA-d) are compared in Table~\ref{tab:hier_mergestrategies}.  
It can be observed: all variants of HCAs are clearly better than a single classifier or ensemble same classifiers in all shots; HCA-d is better than HCA-add and HCA-mul, suggesting that learning-free HCA cannot fully explore the hierarchical information in $\hat{p}_{h}$ and an explicit hierarchical distillation learning is more beneficial.

\begin{table}[!h] \small
    \centering
    \begin{tabular}{c|cccc|c}
    \multirow{2}{*}{Combine}& \multicolumn{4}{c|}{IMDB-WIKI-DIR}&\multirow{2}{*}{SHA}\\
    \cline{2-5}
    &All&Many&Med.&Few \\
    
    \hline
Single CLS&13.58&7.13&	 13.95&	 33.21&58.2\\
Same CLSs&13.42&7.10&14.38&32.22&57.9\\
\hline
Average&          14.85&           7.18&          17.83&          36.24&106.1\\
 HCA-add &12.86&6.98&13.15&30.80&55.9\\
 HCA-mul &12.89&7.00&13.36&30.74&54.7\\
HCA-d &12.70&7.00&13.18&29.94&53.7\\
\hline
    \end{tabular}
    \caption{Comparison of two hierarchical adjustment approaches.}
    \label{tab:hier_mergestrategies} 
\end{table}

\subsection{Analysis of HCA}

\noindent $i$) \textit{coarse classifiers perform ``worse'' due to quantization errors:}  
Fig.~\ref{fig:hier_analy} (a) compares the performance of the classifiers individually versus their quantization error. The coarse classifiers ($1-3$) perform worse individually compared to the vanilla $H$-th classifier, due to the quantization error of representing the entire interval with a single value.

\noindent $ii$) \textit{coarse classifiers provide better range estimation while combining fine classifiers mitigate quantization errors}: In Fig.~\ref{fig:hier_analy} (b), a coarse $h$-th classifier and the finest $H$-th classifier are combined in a coarse-to-fine manner. Specifically, we first get a coarse range prediction from the $h$-th classifier and then select a finer class within this coarse range according to $\hat{p}^H$ predicted by the $H$-th classifier.  It can be observed that merging coarse predictions will significantly decrease the error of the $H$-th classifier, suggesting coarse classifiers provide better range estimation than the finest $H$-th classifier. Meanwhile, selecting a finer class within the coarse range will decrease the bMAE of coarse classifiers ($1\sim 3$), implying that combining fine classifiers can mitigate quantization error in coarse classifiers.

\begin{figure}[!ht]
    \centering
\includegraphics[width=0.45\textwidth]{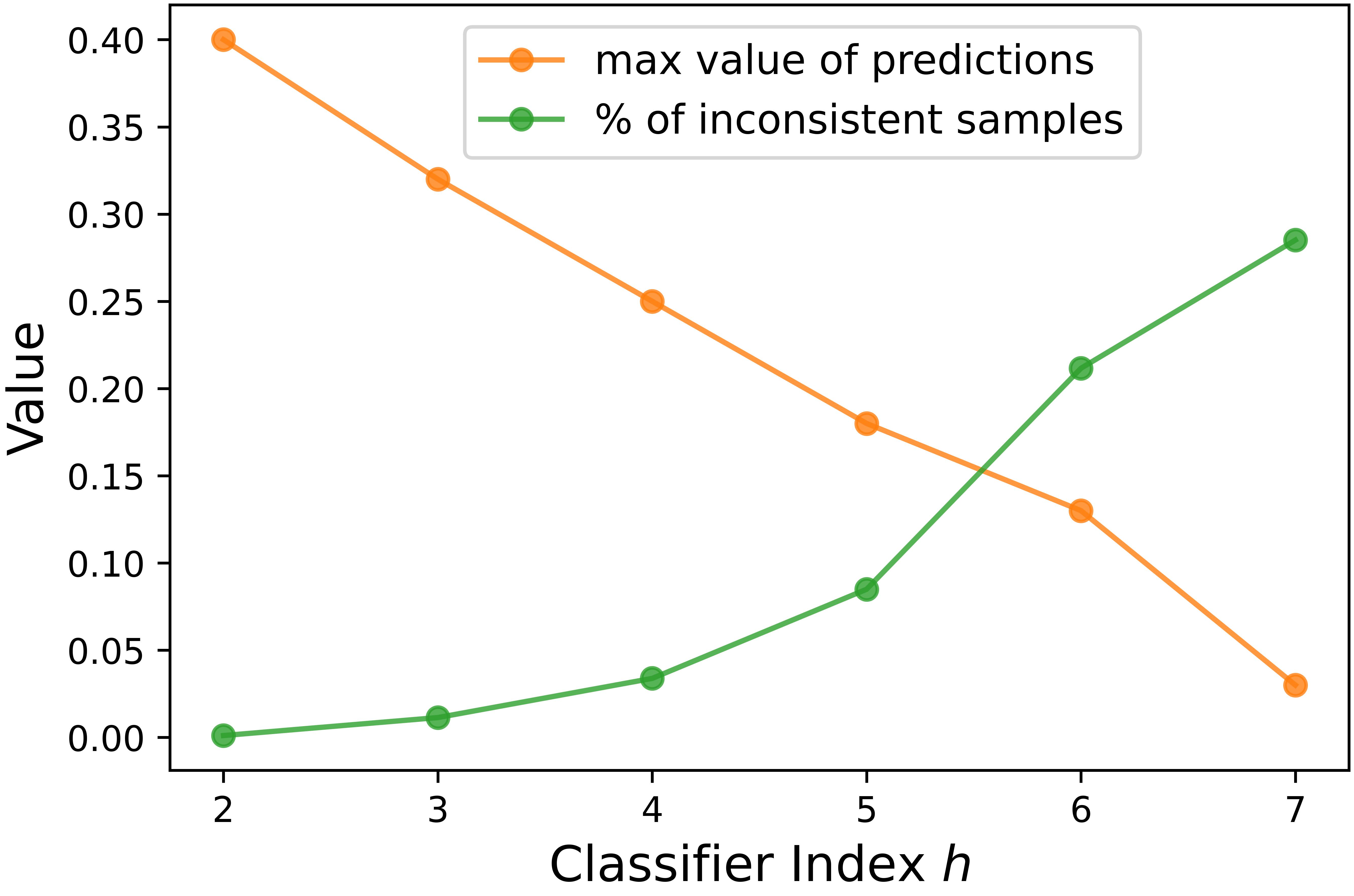}
    \caption{Percentage of inconsistent samples for each hierarchical classifier when downsampled by eq.~\eqref{eq:sum_pool}. The maximum value of hierarchical predictions $\hat{p}_h$ is also visualized.}
    \label{fig:inconsist}
\end{figure}

 \noindent \textit{$iii$) Range-preserving distillation is the key to successful HCA}: In Sec.~\ref{subsec:HCA}, we argue that summation alignment Eq.~\eqref{eq:sum_pool} is not range preserving. Here we experimentally compare summation and range-preserving alignment~\eqref{eq:max_pool} in Table~\ref{tab:hier_align}. It can be observed that using sum alignment harms the performance of HCA in all shots, while the range-preserving alignment can benefit vanilla classification. We further analyze the percentage of samples, which provides inconsistent range estimation when using eq.~\eqref{eq:sum_pool}. Fig.~\ref{fig:inconsist} shows the percentage of inconsistent samples percentage for each classifier head. We can see that when the number of classes increases, the inconsistency increases, which can be explained by the decreased maximum value of prediction $\hat{p}^i$ in finer classifiers. Ideally, if the maximum value of $\hat{p}^h$ is $1$, then the sum of $\hat{p}^h$ will not change the range predicted by $\hat{p}^h$; however, the maximum value of $\hat{p}^h$ will be much less than $1$ in regression by finer classifiers, thus sum operation in eq.~\eqref{eq:sum_pool} cannot ensure consistent ranges cross the hierarchy, as shown in Fig.~\ref{fig:hd_inconsist}.

To justify the influence of second-stage training, we add a ``CLS+GT sup'' baseline, which uses the ground-truth labels to train the classifier $T$. ``CLS+GT sup'' does not present any improvement over the vanilla classification, indicating that hierarchical distillation rather than extra training stages is helpful for imbalanced regression.

\begin{table}[!h] \small
    \centering
    \begin{tabular}{c|cccc|c}
    \hline
    \multirow{2}{*}{Combine}& \multicolumn{4}{c|}{IMDB-WIKI-DIR}&\multirow{2}{*}{SHA}\\
    \cline{2-5}
    &All&Many&Med.&Few \\
    \hline
    CLS&13.58&7.13&	 13.95&	 33.21&58.2\\
    CLS+GT sup&	13.64&7.20&14.94&32.54&57.0\\
    \hline
     HCA sum~\eqref{eq:sum_pool}&27.08&	     15.14&	     38.57&	     55.11&150.3\\
     HCA max~\eqref{eq:max_pool}&12.70&7.00&13.18&29.94&53.7\\
    \hline
    \end{tabular}
    \caption{Comparison of the summation and ranging preserving alignments of hierarchical predictions. ``CLS+GT sup'' denotes using the ground-truth labels rather than hierarchical prediction to supervise the classifier $T$.}
    \label{tab:hier_align} 
\end{table}

\begin{table*}[!ht] \small

    \centering 
    \begin{tabular}{c|ccc|ccc}
    \hline
    \multirow{2}{*}{Configuration}& \multicolumn{3}{c|}{Balanced Subsets}&\multicolumn{3}{c}{Imbalanced Subsets}\\
    \cline{2-7}
    &1000:1000&100:100&10:10&1900:100&190:10&19:1 \\
    
    \hline
    Regression& 6.00$\pm$0.10    & 7.50$\pm$0.04    & 7.56$\pm$0.07    & 6.78$\pm$0.04    & 7.68$\pm$0.05    & 7.74$\pm$0.12\\
    CLS& 6.09$\pm$0.03    & 7.63$\pm$0.05    & 7.61$\pm$0.07    & 6.78$\pm$0.03     & 7.74$\pm$0.12    & 7.90$\pm$0.07\\
    HCA-d& 6.06$\pm$0.04    & 7.53$\pm$0.03    & 7.53$\pm$0.03    & 6.72$\pm$0.04    & 7.54$\pm$0.03    & 7.54$\pm$0.05\\
    \hline
    \end{tabular}
    \caption{Comparison on subsampled balanced and imbalanced subsets of IMDB-WIKI-DIR. Each method is repeated for $5$ times.}
    \label{tab:bal_imbal} 
\end{table*}

\noindent \textit{$iv$) Imbalanced or Insufficient? } 
In imbalanced regression tasks, like age estimation and counting, few-shot samples are always rare ($\leq 20$ samples per class). HCA is helpful in this imbalanced and insufficient scenario, but will it also be helpful in an imbalanced and sufficient dataset where relatively rare classes have plenty of samples? Moreover, is HCA applicable to balanced regression? To verify the influence of imbalance and insufficiency, we resample the IMDB-WIKI-DIR dataset to create these scenarios. Specifically, we first generate balanced subsets with $N$ samples per age and ages ranging from $20$ to $49$. $N$ can be $1000$, $100$ and $10$, covering from sufficient to insufficient scenarios. Then for imbalanced subsets, we take $20\sim34$ ages as many and  $35\sim 49$ ages as few, while keeping the ratios between many to few as $19$. To make a fair comparison, we keep the total sample number the same among balanced and imbalanced subsets, thus having $1900:100$, $190:100$ and $19:1$ imbalanced subsets covering sufficient and insufficient cases. Table~\ref{tab:bal_imbal} presents quantitative results. We can observe that: $i$)HCA does not show significant improvement when the training set is balanced or imbalanced but with sufficient samples ($1900:100$); $ii$) HCA outperforms vanilla classification or regression by a clear margin when the training set is both imbalanced and insufficient.

\subsection{Comparison with SOTA on Regression Tasks}

\noindent \textbf{SHTech Dataset} SHTech~\cite{MCNN_2016_CVPR} is a crowd-counting dataset, which presents severe imbalanced distribution~\cite{xiong2019open,liu2019counting,xiong2022discrete}. It has two subsets, part A and part B. Part A presents crowded scenes captured in arbitrary camera views, while part B presents relatively sparse scenes captured by surveillance cameras. We follow the same network setting as~\cite{xiong2022discrete}, where $100$ logarithm classes are adopted for $C_H$. Mean absolute error (MAE) and rooted mean square error are adopted as evaluation metrics. Both MAE and RMSE are the lower, the better.  Quantitative results are presented in Table~\ref{tab:count}. It can be observed that Hierarchical classification shows the best performance and improves plain classification by a large margin.

\begin{table}[!ht] \small
\centering
\begin{tabular}{l|cc|cc}
\hline
   & \multicolumn{2}{c|}{SHA}   & \multicolumn{2}{c}{SHB}                             \\
  \cline{2-5}
  & MAE$\downarrow$ & RMSE$\downarrow$ & MAE$\downarrow$ & RMSE$\downarrow$   \\
  \hline
CSRNet~\cite{CSRNet_2018_CVPR}     &68.2&115.0  &10.6&16.0       \\
DRCN~\cite{sindagi2020jhu-crowd++}  &64.0&98.4 &8.5&14.4  \\
BL~\cite{ma2019bayesian}  &62.8&101.8 &7.7&12.7  \\
PaDNet~\cite{tian2019padnet}& 59.2& 98.1& 8.1& 12.2\\

MNA~\cite{wan2020modeling}&61.9 &99.6& 7.4& 11.3\\
OT~\cite{wang2020DMCount}&59.7&95.7&7.4&11.8\\
GL~\cite{wan2021generalized}&61.3&95.4&7.3&11.7\\

\hline
Regression~\cite{xiong2022discrete} & 65.4 & 103.3  & 10.7  & 19.5\\
DC-regression~\cite{xiong2022discrete} & 60.7  & 101.0  & 7.1     & \textbf{11.0}\\
\hline

 CLS&58.2&96.7&7.0&11.8\\
HCA-add&55.9	   &92.8&\textbf{6.7}	   &11.4\\
HCA-mul&54.7&91.6&6.8&11.4\\
HCA-d&\textbf{53.7}&\textbf{87.8}&\textbf{6.8}&11.8\\
\hline
\end{tabular}
\caption{Comparison on SHTech dataset~\cite{MCNN_2016_CVPR}. Methods are grouped as density map regression, local count regression and classification approaches.}
\label{tab:count}
\end{table}

\noindent \textbf{IMDB-WIKI-DIR Dataset}
IMDB-WIKI-DIR~\cite{yang2021delving} is a large age estimation dataset, which is an imbalanced subset sampled from IMDB-WIKI~\cite{rothe2018deep}. There are $191509$ training samples, $11022$ validation samples, and $11022$ testing samples. 

We choose three baselines of classification, they are: $i$) vanilla classification, which is $H$-th classifier of HCA; $ii$) classification with label distribution smoothing (LDS)~\cite{yang2021delving}, which re-weight samples with inverse class frequency; $iii$) classification with label distribution smoothing (LDS) and ranksim~\cite{ranksim_icml2022} regularization. ranksim~\cite{ranksim_icml2022} regularizes feature space to have the same ordering as label space. Their HCA counterparts are also included. 

Table~\ref{tab:imdb} presents the quantitative results. From Table~\ref{tab:imdb}, we can observe that: $i$) HCA shows clear improvement in bMAE over naive classification baselines. Specifically, HCA-d can improve all the shots for ``CLS'' and ``CLS+LDS'' baselines, while for strong baseline ``CLS+LDS+ranksim'', since the baseline results are already saturated for the many-shot, there is still a slight trade-off between many and few-shot (many-shot bMAE increases from $6.70$ to $6.88$). $ii$) HCA outperforms its regression baselines and other regression approaches. Noted that Balanced MSE~\cite{ren2022bmse} is a logit adjustment version for regression, it improves the few/medium-shot performances via significantly harming the many-shot (bMAE from $7.32$ to $7.56$), while for HCA-d, many-shot performance is roughly maintained or improved.

\begin{table*}[!h] \small
    \centering
    \begin{tabular}{c|cccc|cccc}
    \hline
    \multirow{2}{*}{Methods}& \multicolumn{4}{c|}{bMAE$\downarrow$} & \multicolumn{4}{c}{MAE$\downarrow$}\\
    \cline{2-9}
    &All&Many&Med.&Few &All&Many&Med.&Few\\
    
    \hline
         Regression~\cite{yang2021delving} &13.92 &7.32 &15.93 &32.78 &8.06 &7.23 &15.12 &26.33\\
      Regression+LDS ~\cite{yang2021delving}&13.37&7.55&13.96&30.92 &8.11&7.47&13.41&23.50\\
      Regression+LDS+ranksim~\cite{ranksim_icml2022} &12.83&7.00&13.28&30.51 &7.56&6.94&12.61&23.43\\ 
      Regression+FDS +ranksim~\cite{ranksim_icml2022} &12.39&6.91&12.82& 29.01&7.35&6.81&11.50&22.75\\ 
    Balanced MSE~\cite{ren2022bmse} &12.66 &7.65 &12.68 &28.14 &8.12 &7.58 &12.27 &23.05\\
    DC-regression~\cite{xiong2022discrete}&14.18&7.30&16.04&34.00&8.05&7.18&15.40&26.48\\
    DC-regression+LDS~\cite{xiong2022discrete}&13.04&8.11&13.62&27.82&8.62&8.04&13.50&\textbf{22.04}\\
    \hline
    CLS&13.58&7.13&	 13.95&	 33.21&7.75&7.04&13.60&25.17\\
    HCA-add &12.86&	      6.98&	     13.15&	     30.80&7.53&	      6.90&	     12.70&	     23.53\\
    HCA-mul&	     12.89&	      7.00&	     13.36&	     30.74&	      7.57&	      6.92&	     12.91&	     23.52\\
    HCA-d &12.70&	      7.00&	     13.18&	     29.94&7.54&	      6.91&	     12.69&	     22.96\\
    \hline
    CLS+LDS&12.85&	      7.31&	     13.40&	     29.54&7.84&	      7.25&	     12.53&	     23.56\\
    HCA-add+LDS &12.64&	      7.15&	     12.83&	     29.47&7.66&	      7.09&	     12.20&	     23.31\\
    HCA-mul+LDS&	     12.68&	      7.18&	     13.03&	     29.42&	      7.70&	      7.11&	     12.35&	     23.34\\
    HCA-d+LDS &12.42&	      7.28&	     12.47&	     28.24&7.77&	      7.21&	     12.25&	     22.43\\
    \hline
    CLS+LDS+ranksim &12.33&	      \textbf{6.70}&	     13.16&	     29.10& \textbf{7.25}&	      \textbf{6.63}&	     12.26&	     22.77\\
    HCA-add+LDS+ranksim &12.15&	      6.77&	     12.09&	     28.80&7.26&	      6.72&	     11.39&	     23.48\\
    HCA-mul+LDS+ranksim&	     12.24&	      6.69&	     12.69&	     29.01&	      7.22&	      6.63&	     11.84&	     23.22\\
    HCA-d+LDS+ranksim &\textbf{11.92}&	      6.88&	     \textbf{11.67}&	     \textbf{27.72}&7.31&	      6.82&	 \textbf{10.99}&	     \textbf{22.04}\\
    \hline
    \end{tabular}
    \caption{Comparison on IMDB-WIKI-DIR Dataset. Methods are grouped as regression, classification approaches.}
    \label{tab:imdb}
\end{table*}

\noindent \textbf{AgeDB-DIR Dataset}
AgeDB-DIR~\cite{yang2021delving} is an imbalanced re-sampled version of AgeDB dataset~\cite{moschoglou2017agedb}. It contains $12208$ training samples, $2140$ validation samples and $2140$ testing samples, with ages ranging from $0$ to $101$. Table~\ref{tab:agedb} presents the quantitative results. HCA approaches show consistent improvement over classification baselines and outperform regression approaches.

\begin{table*}[!h] \small
    \centering
    \begin{tabular}{c|cccc|cccc}
    \hline
    \multirow{2}{*}{Methods}& \multicolumn{4}{c|}{bMAE$\downarrow$} & \multicolumn{4}{c}{MAE$\downarrow$}\\
    \cline{2-9}
    &All&Many&Med.&Few &All&Many&Med.&Few\\
    
    \hline
    Regression~\cite{yang2021delving} & 9.72 & 6.62 & 8.80 & 16.66 & 7.57 & 6.61 & 8.73 & 13.48 \\
      Regression+LDS ~\cite{yang2021delving} &9.12&6.98&8.87&13.66 &7.67&6.98&8.87&10.91\\
    Regression+LDS +ranksim~\cite{ranksim_icml2022} &7.96&\textbf{6.34}&7.84&11.35 &\textbf{6.91}&\textbf{6.34}&7.80&9.92\\    
    Balanced MSE~\cite{ren2022bmse}& 8.97 & 7.65 & 7.43 & 12.65 & 7.78 & 7.65 & 7.45 & 9.99 \\
    DC-regression~\cite{xiong2022discrete}&9.70 & 6.82 & 8.77 & 16.16 & 7.65 & 6.82 & 8.70 & 12.55\\
    DC-regression+LDS~\cite{xiong2022discrete}&9.48 & 7.36 & 9.14 & 14.04 & 8.03 & 7.36 & 9.13 & 11.26\\
    \hline
    CLS&	      9.14&	      6.89&	      8.62&	     14.08&7.58&	      6.89&	      8.51&	     11.60\\
    HCA-add &8.95&	      6.91&	      8.26&	     13.53&7.49&	      6.91&	      8.17&	     11.05\\
    HCA-mul&	      8.97&	      6.93&	      8.35&	     13.52&	      7.52&	      6.93&	      8.25&	     11.10\\
    HCA-d& 8.85&	      6.86&	      8.31&	     13.26&7.45&	      6.86&	      8.22&	     10.90\\
    \hline
    CLS+LDS&	      8.75&	      7.17&	      8.29&	     12.27&7.63&	      7.17&	      8.30&	     10.14\\
    HCA-add+LDS &	      8.40&	      7.22&	      7.83&	     11.18&7.53&	      7.22&	      7.82&	      9.61\\
    HCA-mul+LDS&	      8.54&	      7.25&	      8.02&	     11.49&	      7.60&	      7.25&	      8.02&	      9.70\\
    HCA-d+LDS &	      8.46&	      7.11&	      7.80&	     11.64&7.47&	      7.11&	      7.77&	     10.06\\ 
    \hline
    CLS+LDS+ranksim &	      7.99&	      6.66&	      7.21&	     11.20&6.97&	      6.66&	      7.16&	      9.34\\
    HCA-add+LDS+ranksim &	      \textbf{7.82}&	      6.67&	      \textbf{7.12}&	     \textbf{10.59}&6.94&	      6.67&	      \textbf{7.07}&	      \textbf{9.10}\\
    HCA-mul+LDS+ranksim&	      7.85&	      6.68&	      7.14&	     10.71&	      6.95&	      6.68&	      7.10&	      9.17\\
     HCA-d+LDS+ranksim&	      7.87&	      6.74&	      7.14&	     10.66&7.01&	      6.74&	      7.13&	      9.22\\    
    \hline
    \end{tabular}
    \caption{Comparison on AgeDB-DIR Dataset.}
    \label{tab:agedb}
\end{table*}

\begin{table*}[!ht] \small
    \centering
    \begin{tabular}{c|cccccc}
    \hline
    Methods& MAE$\downarrow$& RMSE$\downarrow$& AbsRel$\downarrow$ & $\delta_1$ $\uparrow$& $\delta_2$ $\uparrow$& $\delta_3$ $\uparrow$\\
    
    \hline
    Regression~\cite{yang2021delving}& 1.004 & 1.486 & \textbf{0.179}  & 0.678& 0.908  & 0.975 \\
    Regression+LDS~\cite{yang2021delving}& 0.968 & 1.387 & 0.188  & 0.672& 0.907  & \textbf{0.976} \\
    Regression+LDS+ranksim~\cite{ranksim_icml2022}& 0.931 & 1.389 & 0.183  & 0.699& 0.905  & 0.969\\

    Balanced MSE~\cite{ren2022bmse}& 0.922 & \textbf{1.279} & 0.219  & 0.695& 0.878  & 0.947\\

    \hline
    CLS& 1.011 & 1.512 & 0.184  & 0.678& 0.906  & 0.958\\
    HCA-add& 0.987 & 1.470 & 0.180  & 0.686& 0.909  & 0.961\\
    HCA-mul& 0.991 & 1.478 & 0.181  & 0.685& 0.909  & 0.960 \\
    HCA-d& 0.987 & 1.475 & 0.181  & 0.689& 0.915  & 0.961\\
    \hline
    CLS+LDS& 0.924 & 1.383 & 0.181  & 0.711& 0.909  & 0.965\\
    HCA-add+LDS& 0.919 & 1.375 & 0.180  & 0.710& 0.910  & 0.965\\
    HCA-mul+LDS& 0.920 & 1.377 & 0.180  & 0.710& 0.910  & 0.965 \\
    HCA-d+LDS& 0.911 & 1.367 & \textbf{0.179}  & 0.714& 0.911  & 0.966\\
    \hline
    CLS+LDS+ranksim& 0.904 & 1.335 & 0.182  & \textbf{0.715}& 0.916  & 0.972\\
    HCA-add+LDS+ranksim& 0.901 & 1.330 & 0.181  & 0.714& \textbf{0.919}  & 0.972\\
    HCA-mul+LDS+ranksim& 0.902 & 1.332 & 0.181  & 0.714& 0.918  & 0.972 \\
    HCA-d+LDS+ranksim& \textbf{0.895} & 1.321 & 0.180  & \textbf{0.715}& \textbf{0.919}  & 0.972 \\
    \hline
    \end{tabular}
    \caption{Comparison on NYUD2-DIR dataset. Methods are
grouped as regression and classification approaches.}
    \label{tab:nyud2}
\end{table*}

\noindent \textbf{NYUDv2-DIR Dataset }
NYUDv2-DIR~\cite{yang2021delving} is an imbalanced version sampled from the NYU Depth Dataset V2~\cite{silberman2012indoor}. The depth values range from $0$ to $10$ meters, which are divided into $100$ logarithm classes for $C_H$. Mean absolute error (MAE), rooted mean square error (RMSE), relative absolute error (RelAbs), $\delta_1$, $\delta_2$ and $\delta_1$ are adopted as evaluation metrics. 
Noted that all classes in NYUDv2-DIR have more than $10^7$ samples, which should be all categorized as many-shot classes according to the criteria in IMDB-WIKI-DIR~\cite{yang2021delving} ($>100$ samples). We report the overall results in Table~\ref{tab:nyud2} and detailed results 
can be found in the supplementary. We can observe that HCA shows improvements to its naive classification baselines and it is also comparable to or better than other regression methods. Noted that the improvement of HCA to CLS in NYUDv2-DIR is small. It is because NYUDv2-DIR is imbalanced but with sufficient samples per class, thus HCA does not improve much. This result is also in accordance with simulated experiments in Table~\ref{tab:bal_imbal}.

\section{Conclusion}
This paper proposes a hierarchical classification adjustment (HCA)
for imbalanced regression. HCA leverages hierarchical class predictions to adjust the vanilla classifiers and improves the regression performance in the whole target space without introducing extra quantization errors. On imbalanced regression tasks including age estimation, crowd counting and depth estimation,  HCA shows superior results to regression or vanilla classification approaches. HCA is extremely helpful in imbalanced and insufficient scenarios; while it is also helpful in balanced and sufficient scenarios.

{\small
\bibliographystyle{ieee_fullname}
\bibliography{reference}
}

\end{document}